\documentclass[11pt]{article}

\usepackage[final]{acl}

\usepackage{times}
\usepackage{latexsym}

\usepackage[T1]{fontenc}

\usepackage[utf8]{inputenc}

\usepackage{microtype}

\usepackage{inconsolata}

\usepackage{graphicx}
\usepackage[utf8]{inputenc} %
\usepackage[T1]{fontenc}    %
\usepackage{url}            %
\usepackage{booktabs}       %
\usepackage{amsfonts}       %
\usepackage{nicefrac}       %
\usepackage{microtype}      %
\usepackage{xcolor}         %

\usepackage{amssymb}
\usepackage{multirow}
\usepackage{amsmath}
\usepackage{graphicx}
\usepackage{comment}
\usepackage{array}
\usepackage{subcaption}
\usepackage{caption}
\usepackage{enumitem}
\usepackage{pifont}
\usepackage[most]{tcolorbox}
\usepackage{wrapfig}

\usepackage{float}
\usepackage{enumitem}
\usepackage{fontawesome}
\usepackage{needspace}

\title{What Prompts Don't Say: Understanding and Managing Underspecification in LLM Prompts}

\author{
    Chenyang Yang\quad
  Yike Shi\quad
  Qianou Ma\quad
  \textbf{Michael Xieyang Liu}\thanks{Now at Google DeepMind.}\quad\\
  \textbf{Christian K\"astner}\quad
  \textbf{Tongshuang Wu}
  \\  
    Carnegie Mellon University
  }

\begin{document}
\maketitle
\begin{abstract}
Prompt underspecification is a common challenge when interacting with LLMs. 
In this paper, we present an in-depth analysis of this problem, showing that while LLMs can often infer unspecified requirements by default (41.1\%), such behavior is fragile:
Under-specified prompts are 2x as likely to regress across model or prompt changes, sometimes with accuracy drops exceeding 20\%.
\footnote{Code and data shared in \url{https://github.com/malusamayo/underspec-analysis}.}
This instability makes it difficult to reliably build LLM applications. 
Moreover, simply specifying all requirements does not consistently help, as models have limited instruction-following ability and requirements can conflict. 
Standard prompt optimizers likewise provide little benefit. 
To address these issues, we propose requirements-aware prompt optimization mechanisms that improve performance by 4.8\% on average over baselines. 
We further advocate for a systematic process of proactive requirements discovery, evaluation, and monitoring to better manage prompt underspecification in practice.
\end{abstract}

\newcommand{\toolname}[0]{\textsc{Orbit}}
\newcommand\circleone{\ding{192}}
\newcommand\circletwo{\ding{193}}
\newcommand\circlethree{\ding{194}}
\newcommand\circlefour{\ding{195}}
\newcommand\circlefive{\ding{196}}
\newcommand\circlesix{\ding{197}}
\newcommand\circleseven{\ding{198}}
\newcommand\circleeight{\ding{199}}

\definecolor{mybgcolor}{HTML}{E6F2F0}    %
\definecolor{myframecolor}{HTML}{FF8552} %
\definecolor{mytitlecolor}{HTML}{2C7873} %

\newcommand{\nbc}[3]{
 {\colorbox{#3}{\bfseries\sffamily\scriptsize\textcolor{white}{#1}}}
 {\textcolor{#3}{\sf\footnotesize$\blacktriangleright$\textit{#2}$\blacktriangleleft$}}
 }

 \newcommand\todo[1]{\nbc{TODO}{#1}{red}}
\newcommand{\cyang}[1]{\nbc{CY}{#1}{teal}}
\newcommand{\swcomment}[1]{\nbc{SW}{#1}{blue}}
\newcommand{\ck}[1]{\nbc{CK}{#1}{orange2}}
\newcommand{\mxl}[1]{\nbc{MS}{#1}{violet}}
\newcommand{\cma}[1]{\nbc{CM}{#1}{orange}}

\newcommand{\delete}[1]{\textcolor{deeporange}{\st{#1}}}
\newcommand{\new}[1]{{#1}}

\newcounter{hyp}
\newcommand{\Hypothesis}[1]{%
  \refstepcounter{hyp}%
  \textbf{Hypothesis~\thehyp}: {#1}\label{hyp:\thehyp}%
}

\newcounter{rq}
\newcommand{\RQ}[1]{%
  \refstepcounter{rq}%
  \textbf{RQ~\therq}: {#1}\label{rq:\therq}%
}

\lstset{
  language=Python,
  keywords={},
  commentstyle=\textbf,
  stringstyle=\text,
  extendedchars=false,
  basicstyle=\ttfamily,
  columns=fullflexible,
  frame=single,
  breaklines=true,
  breakatwhitespace=true,
  breakindent=0\dimen0,
  literate={'}{{\textquotesingle}}1 {`}{{\textasciigrave}}1 {"}{{\textquotedbl}}1,
}

\section{Introduction}

As large language models (LLMs) are improving in capabilities and instruction following~\cite{ouyang2022training}, they are increasingly integrated into commercial applications %
through customized instruction \textit{prompts} that can span thousands of words~\cite[e.g.,][]{ openhands_system_prompt_2025, anthropic2025systemprompts}.
Conveying nuanced intentions to LLMs, however, is inherently difficult.
This issue is less significant for end users, as their prompts are typically one-off and considered successful as long as they yield one satisfactory response throughout their interactions.
For LLM application \textit{developers}, the problem is much more serious, as their prompts need to generalize to many different usage scenarios.

\begin{figure*}[t]
    \centering
    \includegraphics[width=\linewidth]{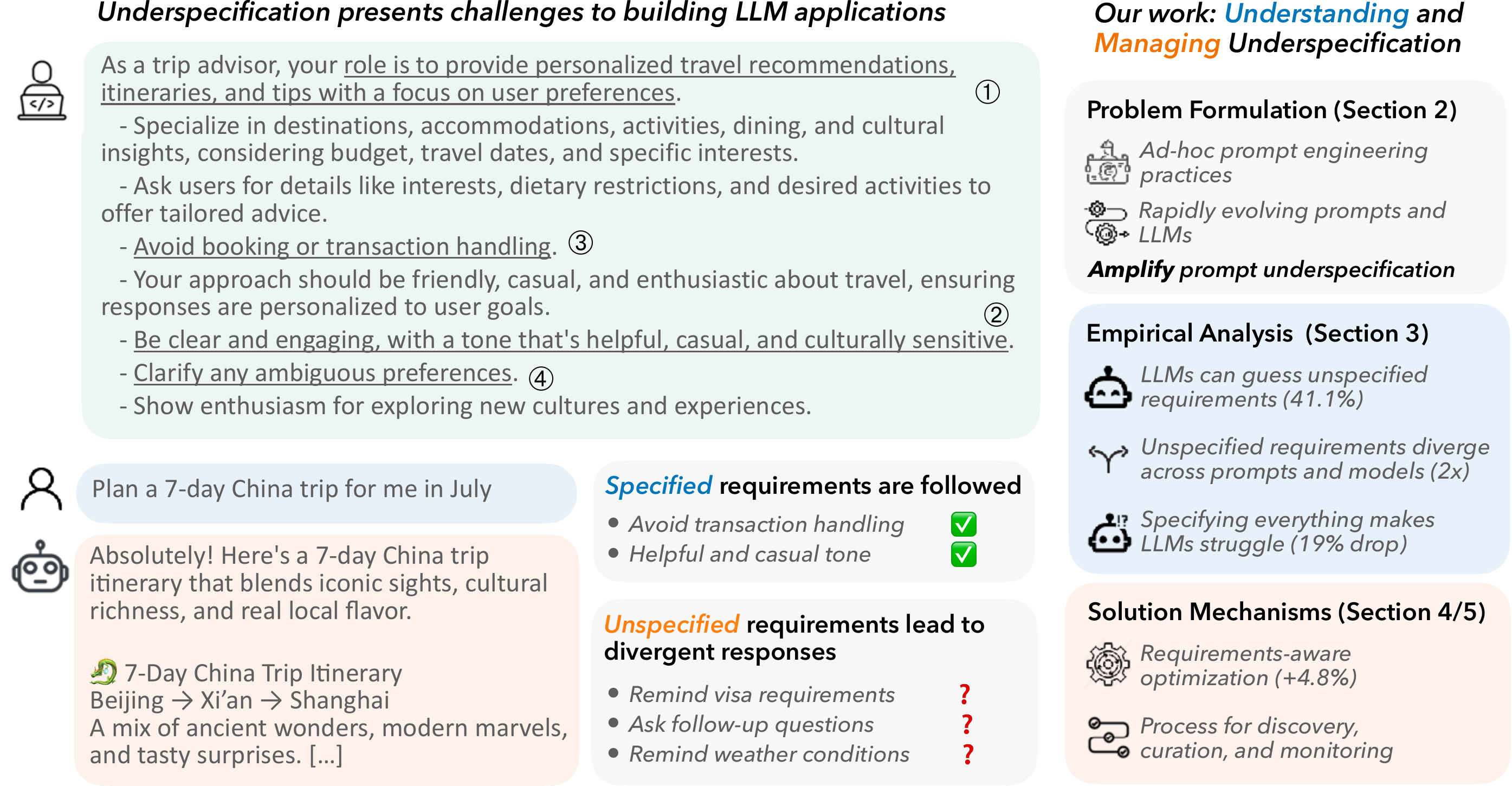}
    \caption{Developers often underspecify prompts and miss user-important requirements, leading to divergent behaviors. 
    We analyze the challenges underspecification presents and propose mechanisms to manage the problem.}
    \label{fig:overview}
\end{figure*}

As an example (Figure~\ref{fig:overview}), consider a developer building an LLM-powered trip advisor:
The detailed prompt may specify the task~\circleone, tone~\circletwo, and certain behaviors such as avoiding transactions~\circlethree{} and clarifying ambiguity~\circlefour, yet it remains \textit{underspecified} in many aspects:
For example, it does not specify whether LLMs should warn about weather, proactively ask follow-up questions, or remind users of visa (entry) requirements. 
If the LLM ends up not satisfying these requirements, it can cause frustrations and failures, such as users booking activities during bad weather, receiving vague recommendations, or facing denied entry due to visa issues, ultimately undermining trust in the LLM-powered applications.
Indeed, failing to mention prerequisites of suggested activities has already caused chaotic experiences in an existing LLM trip advisor~\cite{bernal2024littlefoot}.

We define \textit{underspecification} as \textbf{the omission of essential requirements in a prompt, such that multiple valid but inconsistent behaviors remain possible.}
While it is possible that developers simply do not care enough to specify these behaviors, it is equally -- if not more -- likely that current engineering and evaluation practices make it difficult for developers to identify such underspecification until they have already led to issues in deployment (Section~\ref{sec:problem}). 
Even if developers do not believe these behaviors must be explicitly defined, they may expect models to act consistently along these dimensions to support more stable user mental models of LLM applications,
which is, however, not the case for many unspecified requirements (Figure~\ref{fig:overview}).

In this work, we characterize (Section~\ref{sec:problem}) and empirically analyze (Section~\ref{sec:behaviors}) the problem of prompt underspecification.
To systematically study the problem, we curated a set of diverse requirements across 3 representative tasks, constructed a series of prompts that specify different subsets of the requirements, and evaluated each requirement's satisfaction rate with human-validated LLM-as-a-judge~\cite{zheng2023judging}.
In total, we collected 8.4k data points of LLM+Prompts' aggregated behaviors on diverse requirements.
Our analysis demonstrated that, \emph{while LLMs can indeed often (41.1\%) fill in the underspecification gap, their behaviors are rather inconsistent:}
One version of an LLM may excel in fulfilling an unspecified requirement, but the next version can unexpectedly degrade by more than 20\%.
This will be a problem for continuously developing, deploying, and maintaining LLM-powered applications reliably.

We then introduce \textit{requirements-aware prompt optimization} as a solution strategy to deliberately communicate important requirements to the model, while leaving those already implicitly fulfilled unspecified (Section~\ref{sec:fixing}).
We show that such strategies overcome issues with existing approaches: 
The obvious strategy of simply \textit{specifying all requirements} in the prompt does not work, due to LLMs' limited instruction-following capabilities -- their performance can drop by 19\% as we specify more requirements (Section~\ref{sec:manyreqs}),
and requirement-agnostic prompt optimization only provides limited help since they have no requirement-specific feedback.
We propose and evaluate two \textit{requirement-aware} prompt optimizers:
One to optimize \textit{how} to specify requirements,
and the other to explicitly optimize \textit{what} requirements to specify.
We demonstrate that both strategies work well (+4.8\% accuracy), and the latter can produce shorter prompts (-43\% tokens) that are easier to follow for the model.

Finally, we discuss how to \textit{manage} prompt underspecification when building LLM applications in practice beyond prompt optimization (Section~\ref{sec:managing}):
This includes proactively discovering important requirements, building reliable requirement evaluators, as well as continuously evaluating and monitoring (un-)specified requirements.
We highlight the research opportunities here to support the entire process of managing prompt (under-)specification.

In summary, our work makes the following contribution:
\textbf{(1)} Characterization of the prompt underspecification problem (Section~\ref{sec:problem}), 
\textbf{(2)} an empirical analysis of LLM+Prompt behaviors when underspecified (Section~\ref{sec:behaviors}), 
\textbf{(3)} mitigation mechanisms with \textit{requirements-aware} prompt optimizers (Section~\ref{sec:fixing}), and
\textbf{(4)} a discussion of our vision for managing prompt underspecification (Section~\ref{sec:managing}).

\section{Underspecification is Amplified in Prompting LLMs}
\label{sec:problem}
\label{sec:discover}

While underspecification is a known challenge in traditional software engineering and machine learning pipelines~\cite{kastner2021feature, d2022underspecification}, 
it is significantly amplified in prompting LLMs.
We next elaborate on two key factors behind this amplification: (a) the lack of rigor in prompt engineering practices, and (b) the rapid co-evolution of prompts and LLMs.
A summary of these differences is provided in Table~\ref{tab:spec_comparison}.

\begin{table*}[t]
\centering
\footnotesize
\begin{tabular}{>{\raggedright\arraybackslash}p{1.8cm} >{\raggedright\arraybackslash}p{4cm} >{\raggedright\arraybackslash}p{4cm} >{\raggedright\arraybackslash}p{4.5cm}}
\toprule
\textbf{Aspect} & \textbf{LLM Prompt} & \textbf{Traditional ML Models} & \textbf{Traditional Software} \\
\midrule
\textbf{Specification Method} & Natural language prompts (instructions, examples) & Training data and model architecture and pipeline & Usually natural language and sometimes formal specifications \\
\addlinespace
\textbf{Engineering Practices} & Ad-hoc, trial-and-error prompt iteration~\cite{liang2024prompts, zamfirescu2023johnny} & More structured experimentation pipelines \cite{dmlsbook2022} & Systematic requirements engineering and design processes~\cite{van2009requirements} \\
\addlinespace
\textbf{Artifact Evolution} & Frequent changes with little version control~\cite{tafreshipour2024prompting, liang2024prompts} & Periodical evolution with some version control~\cite{dmlsbook2022} & Evolution often intentionally tracked and limited~\cite{mcconnell1998software,bogart2021and} \\
\addlinespace
\addlinespace
\textbf{Consequences} & Inconsistent and unexpected LLM behaviors (Section~\ref{sec:behaviors}) & Generalization failures, model biases~\cite{d2022underspecification} & 
Software that does not meet customer needs; incorrect behavior in edge cases \cite{SEbook} \\ %
\bottomrule
\end{tabular}

\caption{Compared to traditional ML models and software, LLM prompts are more prone to underspecification, less stable, and evolve more frequently. 
Prompt engineering practices and further amplify these issues.}
\label{tab:spec_comparison}
\end{table*}

\textbf{Ad-hoc prompt engineering amplifies underspecification.}
Prompts are developed with the expectation that not everything needs to be specified -- ideally, LLMs should behave like a human and fill in the gaps with commonsense.
This expectation encourages a \textit{``minimal-specification''} prompt engineering practice:
Developers begin with an initial prompt, observe violations of expected behavior, and iteratively revise through adding more instructions (i.e., specifications)~\cite{zamfirescu2023johnny, liang2024prompts}. 
This trial-and-error process lacks the rigor of traditional requirements engineering~\cite{van2009requirements}, and exposes only a narrow slice of possible behaviors. 

Yet many requirements demand more systematic discovery efforts:
Some are \textit{conditional} (e.g., a \textit{``mentioning visa''} requirement applies only to international travel).
Others are \textit{infrequently violated yet critical} (e.g., failing \textit{``no dangerous activities suggested''} is high-stake even uncommon). 
Still others are \textit{difficult to recognize} (e.g., verifying \textit{``suggested sites are geographically close''}). 
Because such requirements are easily overlooked in ad-hoc iteration, underspecification persists even as prompts appear to perform well on inspected examples.

\textbf{Rapid Prompt-LLM co-evolution amplifies underspecification.}
Prompts are not static artifacts -- they are routinely modified to add new features, address failures, or adapt to updated LLMs~\cite{tafreshipour2024prompting}.
Because their specifications are written in natural language, such revisions are easy and quick to make, encouraging frequent, informal changes.
At the same time, LLMs themselves evolve rapidly, often without clear change logs, and sometimes silently without developers' control~\cite{ma2024my}. 

This dual, fast-moving evolution of both prompts and models introduces continual behavioral drift:
Previously validated behaviors may no longer hold, requiring repeated rediscovery of requirements.
The result is a growing maintenance burden for developers, who must continually re-evaluate prompt behavior to keep up with both axes of change.
In contrast, traditional software change is often carefully managed~\cite{mcconnell1998software,bogart2021and}, and most ML pipelines evolve more slowly.

\section{How do LLM+Prompts Behave when Underspecified?}
\label{sec:behaviors}

\looseness=-1
The difficulty of reliably discovering unspecified requirements raises a natural follow-up question: 
{What happens when such requirements are left out of the prompt?}
While LLM+Prompts are generally known to be unstable, existing work mostly studies their stability on \textit{specified} requirements.
To quantitatively measure LLM+Prompts' behaviors on \textit{unspecified} requirements,
we design an experiment as follows:
We collect a set of 60 plausible requirements across 3 tasks and create human-validated, automated validators for each requirement. 
We then create prompts with subsets of the requirements and measure how well the model's outputs meet the (un-)specified requirements with validators.

Our analysis starts with showing that LLMs can often guess unspecified requirements, but these behaviors are inconsistent (Section~\ref{sec:default-behaviors}) and more likely to degrade with model updates (Section~\ref{sec:instability}). 
We show that an obvious solution of specifying all requirements in a single prompt actually hurts performance, due to LLMs' limited ability to follow long, complex instructions (Section~\ref{sec:manyreqs}).

\subsection{Experiment Setups}
\label{sec:setup}

\textbf{Tasks and data.}
We selected three representative tasks based on Anthropic's report measuring AI usage patterns~\cite{handa2025economic}, covering  different occupational categories (Office, Business, and Computer).
These tasks reflect commercially integrable applications across multiple domains, including software engineering (\texttt{code-explain}), the travel industry (\texttt{trip-advisory}), and e-commerce (\texttt{product-gen}).\footnote{
We analyzed two more tasks, \texttt{health-consulting} and \texttt{lesson-planning}, from different report categories (Physical Sciences, Education), and shared the results in Appendix~\ref{sec:appendix-additional-tasks}.}

\looseness=-1
We re-purposed three existing datasets to run the LLM+Prompts on: Commitpackft~\cite{muennighoff2023octopack} for code explanation, a subset of UltraChat~\cite{ding2023enhancing} for trip advisory, and Amazon ESCI~\cite{reddy2022shopping} for product description generation.
From each dataset, we sampled 200 examples, and split them into training, validation, and test data with 15/35/50 split. More details can be found at Appendix~\ref{sec:appendix-task}--~\ref{sec:appendix-cleaning}.

\textbf{Requirements.}
The key setup of our experiment is to curate a list of plausible requirements for each task we study.
Note that there is no fixed set of requirements for a task, as different developers and application contexts may have different priorities. 
We aim to curate requirements that are plausible and likely shared.
To this end, we explicitly cover various requirements elicitation methods~\cite{van2009requirements} for our curation:

\begin{itemize}[leftmargin=1em, itemsep=1pt, parsep=0pt, topsep=1pt]
    \item \textbf{Existing prompts.} For each task, we collected prompts from the Internet, covering ones provided by Anthropic, Google, and popular GPTs (prompts shared in Appendix~\ref{sec:appendix-curation}). We instructed an LLM (\texttt{gpt-4o}) to extract \textit{specified} requirements from each task prompt. This approach provides requirements that have been incorporated in real-world usage.
    
    \item \textbf{Brainstorming.} For each task, we instructed an LLM (\texttt{gpt-4o}) to analyze potential failure modes and propose requirements. 
    This simulates expert-driven elicitation methods by anticipating system failures~\cite{improvingdomainspecs}.
    
    \item \textbf{Error analysis.} For each task, we ran the curated prompts on the train split with three smaller LLMs. We then instructed an LLM (\texttt{gpt-4o}) to analyze their outputs and suggested missing requirements.
    This simulates error analysis to produce requirements grounded in mistakes.
\end{itemize}

For the elicited requirements, we dropped near duplicates, filtered out the ones that were overly specific, and finally had three independent annotators select the ones they found important for the task. 
We kept the requirements selected by at least one human annotator, with 20 requirements per task.
The final list of requirements covers different categories (content, style, format) and scopes, as visualized in Figure~\ref{fig:req-distribution}.
The full process and curated requirements are shared in Appendix~\ref{sec:appendix-elicitation}.

\looseness=-1
\textbf{Requirement validators.}
For each model output, we validate how well they satisfy each (specified or unspecified) requirement, either using a Python script or an LLM validator~\cite{zheng2023judging}.
We manually validate the LLM validators with a sampled test subset, achieving an average of 95.6\% human-LLM agreement. 
More details can be found in Appendix~\ref{sec:appendix-validation}.

\textbf{Prompts.}
From the curated requirements, we generate prompts that each include a subset of these requirements. 
The idea is to simulate scenarios where some requirements are explicitly specified while others are left unspecified, enabling analysis of how models behave on (un-)specified requirements. 
We use a cyclic design to construct prompts, where each one includes N consecutive requirements (with N set to 10; see Figure~\ref{fig:prompt-gen}). 
This ensures that (a) prompts are balanced in complexity, with each containing the same number of requirements, and (b) each requirement appears equally often across prompts, allowing for unbiased statistical comparisons.

\textbf{Metric.}
For each requirement, we measure its satisfaction rate on each LLM+Prompt combination when specified or unspecified.
We aggregate the results across prompts or models for analysis.

\subsection{LLMs are often able to guess unspecified requirements but lack stability}
\label{sec:default-behaviors}

\textbf{Setups.}
We study three models, one smaller open-sourced model \texttt{Llama-3.3-70B-Instruct},
one big closed-sourced model \texttt{gpt-4o-2024-08-06}, 
and one reasoning model \texttt{o3-mini}.
We calculate the average accuracy and standard deviation for each requirement when specified or unspecified, and compare their distribution.

\begin{figure}
    \centering
    \includegraphics[width=0.9\linewidth]{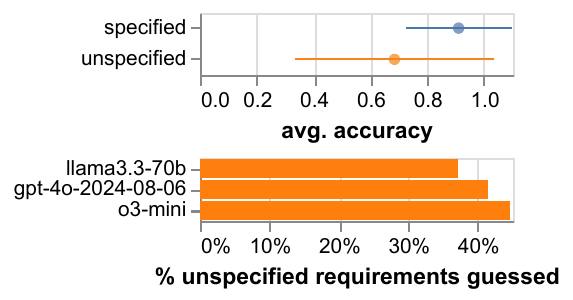}
    \caption{While LLMs perform worse (-22.6\% avg.) when a requirement is unspecified (top), they are often (41.1\% avg.) able to guess unspecified requirements ($\ge0.98$ accuracy), with increased capabilities (bottom). }
    \label{fig:diffs}
\end{figure}

\textbf{Results.}
Unsurprisingly, we generally observe that LLM+Prompts are less likely to implement a requirement when unspecified -- accuracy drops by an average of 22.6\% (and up to 93.1\%) compared to when the requirement is explicitly stated (Figure \ref{fig:diffs}).
Yet, not all requirements are equally affected.
We found that \emph{LLMs are often able to guess unspecified requirements} -- in 41.1\% of cases, they are able to achieve more than 98\% accuracy on unspecified requirements. 
Indeed, for 65\% of all requirements, we found at least one LLM+prompt combination that is able to guess it without explicit specification.

Comparing across models, stronger LLMs are more likely to guess requirements:
\texttt{o3-mini} can guess 44.7\% of unspecified requirements, a 20.2\% increase compared to \texttt{Llama3.3-70b-instruct}, 
possibly due to extra test-time compute to ``infer'' requirements in reasoning.

\looseness=-1
Breaking down the results, we found \emph{LLMs are especially good at guessing {format}-related requirements} (70.7\% vs. 41.1\% on average).
For example, models are often able to \textit{``provide a high-level summary at the beginning''} or \textit{``avoid special characters''} by default.
This is likely because these requirements are more universal and therefore have been built into LLMs natively in post-training.
Meanwhile, LLMs struggle much more with \textit{conditional} requirements (22.9\% vs. 41.1\% ),
as these requirements often specify corner cases that are hard to predict (e.g., \textit{``provide warnings about weather conditions''}).
Somewhat surprisingly, 65.2\% requirements found from existing developer prompts are guessed by LLMs when unspecified.
This indicates that the prompts resulting from existing practices often contain information that might be redundant to LLMs' default behaviors.

While LLMs are often able to guess unspecified requirements, we found them \emph{less robust with unspecified requirements across different prompts}:
Different prompts can guess unspecified requirements completely differently. 
On average, they have a standard deviation of 8.9\%, a more than 2x increase compared to when they are specified.
We observe even a stronger effect when explicitly controlling for requirement conflicts (7.9\% unspecified vs. 0.8\% specification, details in Appendix~\ref{sec:appendix-conflicts}).

\textbf{Implications.}
While LLMs do follow many unspecified requirements, their success depends on how a specific model is post-trained.
Developers seem to struggle with identifying ones that need to be specified.
This justifies a need for automated exploration of what to specify (Section~\ref{sec:optimizer}) and properly managing and testing all relevant requirements, whether specified in the prompt or not (Section~\ref{sec:managing}).

\subsection{LLMs are more likely to regress on unspecified requirements when updated}
\label{sec:instability}

\begin{figure}[t]
  \centering
    \includegraphics[width=0.8\linewidth]{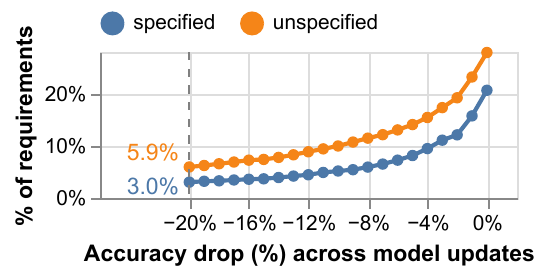}
    \caption{Cumulative distribution of accuracy drop (truncated at 0\%). Prompts regress more on unspecified requirements across model updates, with an almost 2x increase compared to specified requirements.}
    \label{fig:regress}
\end{figure}

\looseness=-1
\textbf{Setups.} 
To analyze model updates, we study six models from \texttt{gpt-4o} and \texttt{llama-3} model families:
Three versions of \texttt{gpt-4o}: \texttt{05-13}, \texttt{08-06}, and \texttt{11-20} for simulating a hidden drift of model versions, and three versions of \texttt{Llama-3-70B-Instruct}: \texttt{Llama-3}, \texttt{Llama-3.1}, and \texttt{Llama-3.3}, simulating intentional model migration and bigger changes.
For each potential update within the same model family, we calculate the accuracy change for each prompt and requirement.

\textbf{Results.}
We found that while the majority of model updates result in stable behaviors (<1\% changes) or improvement, there are still a significant portion (22.9\%) of cases where LLMs regress.
Breaking down the regressions, we found that prompts regress more often on \textit{unspecified} requirements: 5.9\% requirements regress more than 20\% over model updates when they are unspecified -- an almost 2x increase compared to specified requirements (Figure~\ref{fig:regress}).
This is true even for small hidden model updates,
-- e.g., updating \texttt{gpt-4o} from  \texttt{05-13} to \texttt{08-06} makes it 48\% less likely to \textit{``produce skimmable outputs,''} and 14\% less likely to \textit{``mention customer support information''} on average.

\textbf{Implications.}
LLM updates are more likely to improve on specified requirements (through stronger instruction-following) but can hurt unspecified requirements (with different default behaviors).
Regressions of unspecified requirements are both more frequent and far harder to detect.
This makes it necessary to regularly evaluate and monitor known unspecified requirements (Section~\ref{sec:managing}). 
Existing practices of manual inspection (``vibe check'') of a few examples will not be sustainable.

\subsection{LLMs struggle with following many requirements at the same time}
\label{sec:manyreqs}

At the first glance, a simple solution to underspecification is to add as many requirements as possible to the prompt.
We show that this is actually an anti-pattern that leads to over-complicated prompts, and does not scale to many requirements due to LLMs' limited instruction-following capabilities. 

\textbf{Setups.}
We generated additional prompts containing different numbers of requirements ($N$=1, 5, 10, and 19), following the same method described in Section~\ref{sec:setup}.
We study their behaviors on two models, \texttt{Llama-3.3-70B-Instruct} and \texttt{gpt-4o-2024-08-06}, and calculate the average accuracy on $N$ specified and $20-N$ unspecified requirements for each prompt.

\begin{figure}
    \centering
    \includegraphics[width=0.9\linewidth]{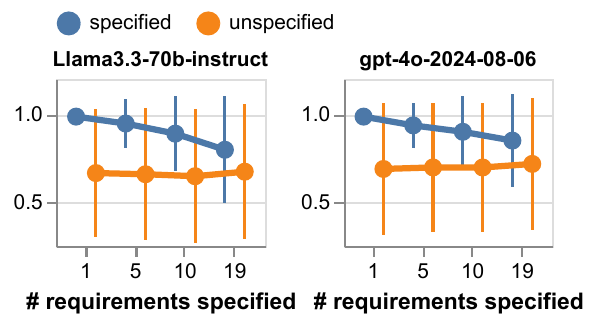}
    \caption{LLMs' average accuracy on \textit{specified} requirements drops with more requirements specified in the prompt, especially for smaller models like \texttt{Llama-3.3-70B-Instruct}.}
    \label{fig:acc-N}
\end{figure}

\textbf{Results.}
First, we found that LLMs are mostly able to follow specified requirements individually,
with an average of 98.7\% accuracy.
This can be thought of as an approximate upper bound on performance, assuming each requirement could be stated in isolation without interference.
Next, we found LLMs' average accuracy starts to drop with more requirements specified (Figure~\ref{fig:acc-N}):
Specifying 19 requirements together yields only an 85.0\% average accuracy for \texttt{gpt-4o}.
Smaller models like \texttt{Llama-3.3-70B-Instruct} struggle even more, with only 79.7\% average accuracy.

Breaking down the results, we found 37.5\% of requirements drop significantly by more than 5\% on average (Figure~\ref{fig:acc-drop-N}).
A few of these requirements suffer from inherent conflicts -- e.g., making product descriptions more skimmable conflicts with other formatting requirements.
However, we also found many cases without obvious conflicts, from \textit{``mentioning availability of transportation options''} (-63.9\% on \texttt{Llama-3.3-70B-Instruct}) to \textit{``use analogies and examples''} (-81.3\% on \texttt{gpt-4o-2024-08-06}).
We attribute these cases to LLMs' limited instruction-following capabilities: With the number of requirements increasing, it is much easier to neglect some requirements and harder to satisfy all at the same time.

\looseness=-1
\textbf{Implications.}
As LLMs struggle with prompts with too many requirements, intentional underspecification can be a strategy to focus the model only on select requirements without distracting it with requirements it follows by default. However, developers currently have no support for intentionally underspecifying their prompts (and are bad at this, see Section \ref{sec:default-behaviors}), which calls for automatically optimizing prompt specification (Section~\ref{sec:fixing}), especially as the number of requirements grows over time.

\section{Requirements-Aware Prompt Optimization}
\label{sec:fixing}

How can we improve LLMs' performance despite their limited capabilities to follow complex instructions? 
Inspired by recent trend on automatically improving prompts~\cite{khattab2023dspy}, 
we test whether existing optimizers can already mitigate this issue, and found that they provide inconsistent improvements.
We then introduce two \textbf{requirement-aware} prompt optimizers:
We first enhance an existing prompt optimizer with requirement-specific evaluators, showing that requirement-specific feedback is valuable.
Inspired by our analysis in Section~\ref{sec:behaviors}, we also explore whether we can optimize \textit{what} requirements to specify in the prompts, removing ones that are distracting or followed by default.
We propose a simple Bayesian prompt optimizer to efficiently search for a good requirement combination.

\begin{table*}[ht]
\centering
\footnotesize
\resizebox{\linewidth}{!}{
\begin{tabular}{lllllll|ll}
\toprule
\multirow{2}{*}{\textbf{Optimizer}} 
& \multicolumn{2}{c}{\texttt{code-explain}} 
& \multicolumn{2}{c}{\texttt{trip-advisory}} 
& \multicolumn{2}{c}{\texttt{product-gen}} 
& \multicolumn{2}{c}{\texttt{website-gen}} \\
\cmidrule(lr){2-3} \cmidrule(lr){4-5} \cmidrule(lr){6-7} \cmidrule(lr){8-9}
& {Acc.} & {\#Tokens} 
& {Acc.} & {\#Tokens} 
& {Acc.} & {\#Tokens} 
& {Acc.} & {\#Tokens} \\
\midrule
- (Original)     & $0.754\pm0.021$ & $342\pm4$ & $0.803\pm0.024$  & $299\pm4$ & $0.835\pm0.026$ & $303\pm4$  & 0.441 & 431 \\
OpenAI   & ${0.774}\pm0.049$ & $765\pm154$ & $0.798\pm0.066$ & $1233\pm238$ & $0.845\pm0.031$ & $664\pm220$ & 0.504 & 1158\\
COPRO   & $0.804\pm0.053$ & $351\pm54$  & ${0.785}\pm0.033$ & $234\pm56$ & $0.868\pm0.041$ & $207\pm44$  & -  & -\\
COPRO-R  & $\textbf{0.842}\pm0.049$ & $337\pm55$  & $\textbf{0.811}\pm0.022$ & $281\pm40$ & $0.913\pm0.035$ & $220\pm50$  & \textbf{0.535} & 654 \\
Bayesian & ${0.773}\pm0.025$&  ${187}\pm26$ &  $\textbf{0.810}\pm0.040$& ${170}\pm20$ & $\textbf{0.922}\pm0.028$ & ${147}\pm34$ & {0.474} & 121\\
\bottomrule
\end{tabular}
}
\caption{Prompt optimization results on 60 different prompts. We found both requirement-aware optimizers (COPRO-R and Bayesian) can consistently improve prompt performance (+4.8\% on average), with the Bayesian optimizer reducing token usage by 41 - 45\%.
The trend holds for real-world prompts in agentic coding context. }
\label{tab:optimizers}
\end{table*}

\subsection{Existing prompt optimizers do not improve performance consistently}
We first explore whether we can improve prompts with existing prompt optimizers. We use two off-the-shelf prompt optimizers here:
   \textbf{(1)} OpenAI's prompt optimizer~\cite{openai2025promptgen}: This is a ``static'' prompt optimizer that takes in a prompt and tries to improve it without any model execution feedback.
   \textbf{(2)} DSPy's COPRO optimizer~\cite{khattab2023dspy}: This represents a ``dynamic'' prompt optimizer that iteratively proposes and explores new prompts and finds ones with higher performance. 
    To guide the optimization, we use an LLM evaluator that scores outputs from 1 to 10, based on how well they adhere to the input prompts.

\textbf{Setups.}
We apply all optimizers to prompts with the most requirements (N=19) and an LLM with weaker instruction-following capabilities \texttt{Llama3.3-70b-instruct}. 
We perform prompt optimization on the train split (n=30) of each task dataset, and evaluate the results on the test split.
All dynamic prompt optimizers are given a budget of 9 prompts to explore.
We measure average requirement accuracy and prompt token usage.

\textbf{Results.}
Overall, we do not observe consistent improvements from the optimizers (Table~\ref{tab:optimizers}, OpenAI and COPRO).
While they provide small improvements on two tasks (+2.8\% on average), they also drop prompt performance on the \texttt{trip-advisory} task (-1.1\% on average).

\subsection{Designing requirement-aware prompt optimizers}
\label{sec:optimizer}
Next, we explore whether we can leverage requirements to help with prompt optimization: 

\textbf{COPRO-R.}
We first enhance COPRO with requirement-specific validators to guide its optimization, providing average accuracy of \textit{all} requirements (whether they are specified or not in the prompt) as the optimization metric.
We expect this helps optimizers obtain more accurate feedback and generate prompts that consider different requirements thoroughly.

\textbf{Bayesian.}
We then explore optimizing \textit{what} requirements to specify in a prompt.
This is from our observations that many requirements are actually redundant, followed by default, and do not need to be specified (Section~\ref{sec:default-behaviors}).
To explore an exponential number of requirement combinations, we propose a simple Bayesian prompt optimizer:
We model each requirement as a binary hyperparameter (specified vs.\ unspecified), with a configuration \( r = (r_1, r_2, \ldots, r_n) \in \{0,1\}^n \) indicating what requirements to specify.
Given a performance function \( f(r) \), the optimizer aims to solve the objective:
\(r^* = \arg\max_{r \in \{0,1\}^n} f(r)\).

We define \( f(r) \) as average requirement accuracy for our experiment, and leverage a classic Bayesian optimization algorithm, Tree-structured Parzen Estimator~\cite{bergstra2011algorithms}, to efficiently find a good \( r \) in a small number of trials.

\textbf{Results.}
We found that both requirement-aware optimizers can consistently improve prompt performance (Table~\ref{tab:optimizers}).
The simple Bayesian optimizer improves prompt performance by 3.8\%, while reducing token usage by 41 - 45\%. 
Inspecting the produced prompts, we found that global, format, and developer-written requirements are more likely to be dropped (56.5\%, 77.0\%, 57.9\% respectively vs. 52.8\% avg.), which aligns well with our observations in Section~\ref{sec:default-behaviors} that these requirements are more likely to be guessed by an LLM.

Pairing requirement-specific validators with existing optimizers, we found COPRO-R can also improve performance by 5.8\%.
Inspecting the produced prompts (Figure~\ref{fig:optimized-prompt}), we found they reorder requirements in a more logical structure, merge related requirements together, and sometimes drop requirements from the list.
These together produce a better way to specify a longer list of requirements.
In contrast, the Bayesian optimizer offers a complementary strength to improve performance by deciding what \textit{subset} of requirements to specify.

In addition, we found our optimizers generalize well to real-world contexts: 
On the agentic coding task of \texttt{website-gen} (see Appendix~\ref{sec:appendix-case-study} for setup details), we found both optimizers are able to significantly improve baseline prompt shared in the Cursor community, with the Bayesian optimizer reducing token usage by 71.9\%.

\section{Towards Managing Prompt Underspecification}
\label{sec:managing}

\looseness=-1
While requirement-aware prompt optimizers are effective, their success depends on a more rigorous, end-to-end process for building LLM-powered applications. Drawing on insights from software engineering~\cite{SEbook}, we outline a structured approach to managing underspecification -- where prompt developers can \textit{discover} key task requirements, \textit{curate} them to reflect real needs, and continuously \textit{monitor} for drift.

\textbf{Elicit requirements for comprehensive task representation, by increasing requirement discoverability.}
As emphasized in the software engineering literature~\cite{van2009requirements}, requirement elicitation is foundational for application development. 
However, as discussed in Section~\ref{sec:discover}, identifying task-specific requirements remains nontrivial. 
While our experiments take early steps toward automating this process (Section~\ref{sec:setup}), the broader challenge is to \emph{maximize} requirement discoverability. 
Future work may explore using synthetic data generation~\cite{zhao2024self} to surface edge cases or probe specific requirements.
We may also explore more traditional requirement engineering approaches, including top-down brainstorming~\cite{weaver}, bottom-up data-driven analysis~\cite{zeng2025evaltree}, 
or structrued safety engineering approaches that anticipate problems before they occur~\cite{hong2025hazard}.

\textbf{Select requirements that matter, via continuous monitoring on the full requirement set.}
LLM behaviors drift over time~\cite{chen2024chatgpt} and they have different capabilities to follow requirements that are specified and guess ones that are not (Section~\ref{sec:default-behaviors}).
Deciding what requirements to specify and how to specify them (Section~\ref{sec:fixing}), therefore, is largely model-dependent.
To support model migrations, we recommend background validation of all tracked requirements. 
This allows detection of drift and allows developers or optimizers to re-select which requirements to explicitly specify when switching models, from a stable overarching set. 
This requires running validations for each new prompt or model and alerting developers of significant changes.
We could further distribute requirements across sub-modules in workflows or multi-agent systems~\cite{zhou2025multi} to enabling more local monitoring and optimization.

\textbf{Curate requirements to be more aligned with developer needs, by validating the validators.}
Both optimization and monitoring rely on robust requirement validators. 
While validators can be separately tuned (e.g., we performed manual validation), a more rigorous approach is to close the loop between requirements and their validators. 
If a validator gives unstable (e.g., low-confidence or inconsistent) outputs -- especially when compared to human annotations -- it likely signals ambiguity or misalignment in the requirement itself.
To do so, future work can invest in meta evaluation and alignment of LLM-as-judge~\cite{shankar2024validates}, strategically have developers review (intentionally curated) validation results, and develop strategies that can refine the requirements.

\section{Related Work}

\looseness=-1
\textbf{Instruction following capabilities of LLMs.}
Much research has investigated LLMs' instruction-following capabilities, from building datasets~\cite{zhou2023instruction, qin2024infobench}, training better models~\cite{sanh2021multitask, ouyang2022training}, to optimizing for task-specific instruction-following capabilities~\cite{zhao2024self}.
Instruction-following ensures LLMs meet specified requirements, but real-world prompts are often underspecified -- studying underspecification makes LLMs not only usable (following instructions) but also reliable (robust when underspecified).

\looseness=-1
\textbf{Empirical analysis of LLM behaviors.}
Lots of work has empirically analyzed the behaviors of LLMs:
They found that LLMs are sensitive to prompt design~\cite{sclar2023quantifying, cao2024worst}, that different LLMs exhibit different behaviors~\cite{dunlap2024vibecheck, sun2025idiosyncrasies}, and that LLM updates can often trigger unexpected performance regression~\cite{chen2024chatgpt, ma2024my}.
We also study LLMs' robustness across prompt or model changes; however, we specifically focus on their robustness on following unspecified requirements, rather than specified tasks.

\textbf{Ambiguity resolution when interacting with LLMs.}
Ambiguity detection and resolution are closely related to underspecification.
For interactive applications, asking clarifying questions can be used as a fallback mechanism to resolve where user prompts are underspecified~\cite{zhang2023clarify, zhang2024clamber}.
However, studies found that LLMs often struggle to detect ambiguity in user queries~\cite{vijayvargiya2025interactive, ma2024engineerpromptstraininghumans,laban2025llms}, and much work has tried to improve LLMs' abilities to handle ambiguity~\cite{kim2024aligning}.

\textbf{Prompt optimization.}
Prompt optimization has been extensively studied to adapt the prompt to improve LLM performance~\cite{khattab2023dspy, yuksekgonul2024textgrad, li2024learning}.
They require a clear metric to guide the optimization.
Our work complements their work by focusing on \textit{identifying what metrics to optimize} through properly discovering and operationalizing task requirements to construct a multi-faceted optimization objective. 

\section{Conclusion}
Prompt underspecification is common and challenging: 
LLMs can infer missing requirements but behave inconsistently across prompts and models, while fully specified requirements overwhelm them. 
We show that requirements-aware optimizers make prompts easier to follow and argue that systematic discovery, evaluation, and monitoring can help effectively manage underspecification.

\section*{Limitations}
\label{sec:limitations}
\textbf{Scale of experiments.} 
Our experiments are conducted on a relatively small number of requirements (n=60) and synthetic prompts (n=240). 
This setup allows us to curate high-quality, human-validated, realistic requirement–validator pairs and study underspecification systematically, but it limits the breadth of our analysis. 
To scale up, future work will need to curate a larger set of realistic requirements and validators -- note that high-quality validators usually require manual validation~\cite{shankar2024validates}. They will also need more efficient ways (our experiments involved over 1.5 million LLM calls, see Appendix~\ref{sec:appendix-compute}) to produce fine-grained requirement-specific evaluation results while keeping the results reliable.

\looseness=-1
\textbf{Evaluation methodology.} 
We rely on LLM validators to scale up the evaluation. 
While we manually validate their reliability and ensure a small error rate, LLM validators can exhibit preference biases toward models from the same family. 
This, however, does not affect our main results, which do not compare across model families (e.g., \textit{``model updates regress more on unspecified requirements''}). 

\textbf{Connection to practices.} 
Our primary goal in this work is to first establish an empirical understanding of the underspecification problem. 
Accordingly, we do not include a user study or direct integration with developer workflows. 
This raises the possibility that the observed patterns may not transfer cleanly into practice. 
Future work should evaluate how well these findings generalize when embedded in real-world processes, constraints, and decision-making contexts of prompt development.

\section*{Acknowledgments}
This work is supported by the OpenAI Research Credit program, the Amazon AI Research gift fund, and the Gemma Academic Program GCP Credit Award.
We thank Maarten Sap for generously providing us with additional computational resources to run the experiments.
We thank Xinran Zhao, Vijay Viswanathan, Jessie Mindel, Jenny T. Liang, Nadia Nahar, Manisha Mukherjee, Vasu Vikram, and Anjiang Wei for their feedback on this work.

\bibliography{main}

\appendix

\section{Details on Experiments Setups}

\subsection{Task descriptions}
\label{sec:appendix-task}
We selected three tasks based on Anthropic's report measuring AI usage patterns~\cite{handa2025economic}: 
\begin{itemize}[noitemsep, topsep=0pt]
    \item \texttt{trip-advisory}: Provide personalized travel recommendations, itineraries, and tips.
    \item \texttt{product-gen}: Write engaging product descriptions from the provided product details.
    \item \texttt{code-explain}: Explain a code snippet for learning purposes.
\end{itemize}

\subsection{Process of data sampling and cleaning}
\label{sec:appendix-cleaning}
We re-purposed three existing datasets to run the LLM+Prompts on: Commitpackft~\cite{muennighoff2023octopack} for code explanation (MIT license), a subset of UltraChat~\cite{ding2023enhancing} for trip advisory (MIT License), and Amazon ESCI~\cite{reddy2022shopping} for product description generation (Apache-2.0 License).

\begin{itemize}[noitemsep, topsep=0pt]
 \item For Commitpackft, we take the python split\footnote{\url{https://huggingface.co/datasets/bigcode/commitpackft}}, we keep all examples with more than 90 lines of code (n=357), and then take the first 200 examples.

 \item For UltraChat, we take a travel-related subset\footnote{\url{https://huggingface.co/datasets/soniawmeyer/travel-conversations-finetuning}} and keep the first 800 examples.
We then use \texttt{gpt-4o-mini} to label if each query is asking for travel recommendations, itineraries, and tips, and filter out the queries that are not (n=248 remains). We then take the first 200 examples.

 \item For Amazon ESCI, we take the test split\footnote{\url{https://huggingface.co/datasets/tasksource/esci}}, filter out the examples that are not based in US and not in English, and remove duplicated products.
We then sample 200 examples from the dataset.
\end{itemize}

From each dataset, we split it into training, validation, and test data with 15/35/50 split. 
All our evaluation results are reported based on the test split.
All datasets are available in our shared code repository.

\subsection{Process of requirements curation}
\label{sec:appendix-elicitation}
From each task, we curated an initial list of requirements through the three different sources described in Section~\ref{sec:behaviors}.
We found existing prompts provided by Anthropic, Google and GPTs\footnote{\url{https://docs.anthropic.com/en/prompt-library/code-clarifier},\\ \url{https://github.com/google-marketing-solutions/feedgen/blob/main/img/config.png}, \\\url{https://github.com/yourzxb/GPTs/blob/main/17/Trip Advisor.md}} (Figure~\ref{fig:all_prompt_1}).
We kept all requirements that are curated from existing prompt, as they already approved by some human developers.

\begin{figure}
    \centering
    \includegraphics[width=\linewidth]{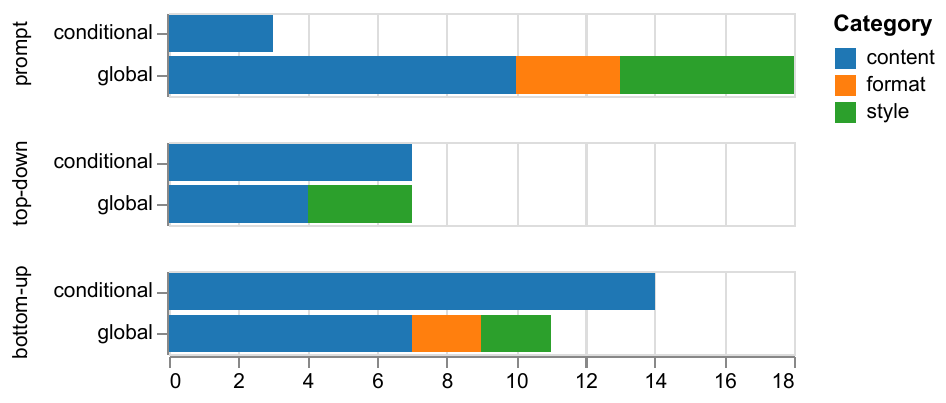}
    \caption{We gather 60 requirements for our analysis. The majority of requirements come from bottom-up error analysis (41.7\%), followed by existing prompts (35\%), and top-down brainstorming (23.3\%). 
    Most requirements specify content-related constraints (75\%), followed by style (16.7\%) and format (8.3\%).
    Most requirements are global and apply to all examples (60\%), while 40\% are conditional requirements.
    We found that existing prompts rarely consider conditional requirements (only 14.3\%).
    }
    \label{fig:req-distribution}
\end{figure}

\begin{figure}[ht]
\small
\begin{lstlisting}
Instruction: Review the provided list of requirements and select up to 10 that you believe are the most important for the task. 
For each selected requirement, include a brief justification explaining why it is important.

Task description: Explain the code snippet.

Sample inputs: 
```python
from sqlalchemy.exc import IntegrityError

from ggrc import db
from ggrc import models
from integration.ggrc import TestCase


class TestCAD(TestCase):
  """Tests for basic functionality of cad model."""

  def test_setting_reserved_words(self):
    """Test setting any of the existing attribute names."""

    with self.assertRaises(ValueError):
      cad = models.CustomAttributeDefinition()
      cad.definition_type = "Section"
      cad.title = "title"

    with self.assertRaises(ValueError):
      cad = models.CustomAttributeDefinition()
      cad.title = "title"
      cad.definition_type = "Section"
```
\end{lstlisting}
\caption{Sample annotator instruction during requirement curation.}
\label{fig:annotator-prompt}
\end{figure}

We use \texttt{gpt-4o} to elicit requirements through brainstorming and error analysis.
The error analysis is conducted on the responses generated from a set of smaller LLMs (\texttt{gpt-4o-mini}, \texttt{gemini-1.5-flash}, \texttt{llama3.2-11b}).
For elicited requirements, we use \texttt{text-embedding-ada-002} to generate embeddings of each requirement and remove ones with high cosine similarity ($>0.9$) to other existing requirements incrementally.
After this step, we curate 38, 39, and 40 requirements for \texttt{trip-advisory}, \texttt{product-gen}, and \texttt{code-explain} tasks respectively.

We then filtered out the requirements that are overly specific (e.g., \textit{``The output must explain how the product's features enhance the karaoke experience for the targeted age group.''}), and finally had three independent annotators select important requirements. 
We recruited the annotators from our contacts and ensured that they all have sufficient background knowledge for the task they need to annotate (e.g., all three annotators are in a computer science PhD program for the code explanation task). 
The annotators have access to a short annotation rubric, including an instruction that asks them to mark up important requirements for the task and add justifications when needed, a task description, and example inputs to ground their annotations (Figure~\ref{fig:annotator-prompt}).
The annotation process takes up to 20 minutes, and the annotators are not compensated for their time.
We verbally communicated to the annotators that their selected requirements will be used for later experiments.

We kept the requirements selected by at least one human annotator in the end, with 20 requirements for each and a total of 60 requirements as in Section~\ref{sec:appendix-requirements}.
Note that different annotators may diverge in the requirements they selected, and we intentionally kept the diversity from the process, as it is natural that different stakeholders may have different priorities~\cite{van2009requirements}.
The final list of requirements covers different categories (content, style, format), different scopes (global, conditional), as visualized in Figure~\ref{fig:req-distribution}.

We provide all requirements used in our experiments in the shared code repository along with their validators. 

\begin{figure}[t]
\small
\begin{lstlisting}
Your task is to take the code snippet provided and explain it in simple, easy-to-understand language. Break down the code’s functionality, purpose, and key components. Use analogies, examples, and plain terms to make the explanation accessible to someone with minimal coding knowledge. Avoid using technical jargon unless absolutely necessary, and provide clear explanations for any jargon used. The goal is to help the reader understand what the code does and how it works at a high level.
\end{lstlisting}
\begin{lstlisting}
You are a leading digital marketer working for a top retail organization. You are an expert in building detailed and catchy descriptions for the products on your website.

Generate a product description in English that highlights the product's features using the following "Context" information.
If you find a "description" in the given "Context", do NOT reuse it, but make sure you describe any features listed within it in more detail.
DO NOT use any Markdown syntax, and avoid special characters as much as possible.
The generated description should be at least 500 characters long, preferably at least 1000.
\end{lstlisting}
\begin{lstlisting}
As a trip advisor, your role is to provide personalized travel recommendations, itineraries, and tips with a focus on user preferences. 
   - Specialize in destinations, accommodations, activities, dining, and cultural insights, considering budget, travel dates, and specific interests. 
   - Ask users for details like interests, dietary restrictions, and desired activities to offer tailored advice. 
   - Avoid booking or transaction handling. 
   - Your approach should be friendly, casual, and enthusiastic about travel, ensuring responses are personalized to user goals. 
   - Be clear and engaging, with a tone that's helpful, casual, and culturally sensitive. 
   - Clarify any ambiguous preferences. 
   - Show enthusiasm for exploring new cultures and experiences.
\end{lstlisting}
\caption{Existing prompts for the three studied tasks. We extracted a subset of requirements from these prompts and constructed a set of synthetic prompts with the prompt templates in Appendix~\ref{sec:appendix-prompts}}
\label{fig:all_prompt_1}
\end{figure}

\clearpage

\subsection{Prompts for requirements elicitation and validation}
\label{sec:appendix-curation}

\begin{figure}[ht]
\small
\begin{lstlisting}
You are an experienced requirements engineer. Your goal is to brainstorm a list of requirements that specify desired LLM behaviors for the given task.
These requirements should identify behaviors that, if omitted, would likely frustrate or annoy users -- such as forgetting to surface important reminders, warnings, or common-sense.

These requirements should be consistent with each other without contradictions and complementary to existing requirements.
    
Guidelines:
- Each requirement should test exactly ONE requirement
- Requirements should be easily verifiable, almost as if writing a Boolean condition in Python. They should be testable with Python code or an LLM itself (no human judgment or external sources needed).
- Requirements should not be overly general (i.e. they should not be universal requirements that might apply to any reasonable reasponse)
- Requirements should be generally applicable for responses to that task, not referring to any specific response
- Avoid unrealistic edge cases - focus on plausible failures that could occur even in aligned or well-trained LLMs.
- Focus only on objective, measurable requirements
- Use concise and unambiguous language
- Never generate similar requirements to the existing requirements
\end{lstlisting}
\caption{Prompts for requirement elicitation - Brainstorming.}
\label{fig:elicitation_prompts_1}
\end{figure}

\begin{figure}[t]
\small
\begin{lstlisting}
You are an experienced requirements engineer. Your goal is to extract a list of atomic requirements that specify desired LLM behaviors for the given task.

You will be presented with a model input and several model outputs from different models. First, provide a detailed analysis critiquing the model outputs.
Then, based on the analysis, suggest a list of atomic requirements that specify desired LLM behaviors for the given task.
These requirements should be consistent with each other without contradictions and complementary to existing requirements.

Guidelines:
- Each requirement should test exactly ONE requirement
- Requirements should be easily verifiable, almost as if writing a Boolean condition in Python
- Requirements should not be overly general (i.e. they should not be universal requirements that might apply to any tasks)
- Requirements should be generally applicable for responses to that task, not referring to any specific input examples
- Focus only on objective, measurable requirements
- Use concise and unambiguous language
- The requirements should be consistent with each other without contradictions
- The requirements should not overlap with existing requirements

Here are some bad requirements:
- The output should be interesting. - This is subjective
- The output should provide examples in fewer than 280 characters. - This overloads multiple aspects
- The output should be helpful and harmless. - This is overly general

Here are some good atomic requirements:
- The output should provide examples.
- The output should be fewer than 280 characters.
- The output should contain at least 3 references.
\end{lstlisting}
\caption{Prompts for requirement elicitation - Error analysis.}
\label{fig:elicitation_prompts_2}
\end{figure}

\begin{figure}[t]
\small
\begin{lstlisting}
You are a reviewer who is evaluating whether a model output satisfies the given requirement. Given a task description, examples, and requirement, draft a step-by-step evaluation plan for the requirement. 

Follow the guideline below:
- The evaluation plan should be a step-by-step process to evaluate if the requirement is met.
- The evaluation plan should be concise and clear, and lead to a final judgment on whether the model output meets the requirement. 
- When requirements are conditional (e.g., indicated by "if applicable"), the evaluation plan should include a first step to check if the requirement is applicable to an example input.
- The evaluation plan should only include the steps to evaluate the requirement, and not include any additional feedback or suggestions, or steps to evaluate other related requirements.

Examples
---
Requirement: The explanation should provide examples of how to instantiate and use key classes, if applicable.
Evaluation Plan:
1. Identify the key classes in the given code snippet by examining the code structure and class definitions. If there are no key classes, this requirement is not applicable.
2. Check that the explanation clearly highlights which classes are considered "key" for this snippet (for example, any classes that define core functionality or are central to the code's purpose).
3. Verify that the explanation includes concrete examples showing how to instantiate the identified key classes.
4. Finally, assess whether the explanation meets the requirement by providing sufficient instantiation and usage examples that a user could follow.
\end{lstlisting}

\begin{lstlisting}
You are a reviewer who is evaluating whether a model output satisfies the given requirement. Given a task description, examples, and requirement, write a Python function to evaluate the requirement.
    
The Python function `evaluation_function` takes task_description, model_input, and model_output as input arguments and returns a boolean value indicating whether the requirement is met.
\end{lstlisting}

\begin{lstlisting}
You are a reviewer who is evaluating whether a model output satisfies the given requirement. 
    
Given a task description, model input, model output, a requirement and its step-by-step evaluation plan, execute the evaluation plan to evaluate if the model output meets the requirement. If the requirement is not applicable, return True for meets_requirement.
\end{lstlisting}
\caption{Prompts for requirement evaluation: Planning (top and middle) and execution (bottom).}
\label{eval_prompts}
\end{figure}

\clearpage

\subsection{Complete list of curated requirements}
\label{sec:appendix-requirements}

\begin{figure*}[h]
\begin{tcolorbox}[title=Curated requirements for \texttt{trip-advisory}, colback=blue!5!white, colframe=blue!75!black, fonttitle=\bfseries, label={box:trip-advisory-requirements}]
\begin{itemize}[leftmargin=1.5em]
\item The output should consider factors such as budget, travel dates, and specific interests.
\item The output should ask users for details like interests, dietary restrictions, and desired activities.
\item The output should not include booking or transaction handling.
\item The output should be friendly, casual, and enthusiastic about travel.
\item The output should be personalized to user goals and preferences.
\item The output should be clear and engaging.
\item The output should be culturally sensitive.
\item The output should clarify any ambiguous preferences.
\item The output should show enthusiasm for exploring new cultures and experiences.
\item The output should ensure activities are age-appropriate if age preferences are specified by the user.
\item The output should highlight any visa or entry requirements specific to the suggested destinations.
\item The output should provide warnings about weather conditions that might affect accessibility to certain activities during the user's planned travel dates.
\item The output should provide follow-up questions to solicit user preferences if they are not initially provided.
\item The output should clarify the geographic context if the location is ambiguous.
\item The output should specify if transportation options are seasonal or subject to availability.
\item The output should include suggestions for public transport or alternative travel methods.
\item The output should correctly identify and focus on sites located within the specified geographic area.
\item The output should provide references to local regulations or park rules, when applicable.
\item The output should contain a section explicitly stating safety guidelines specific to solo traveling.
\item The output should include tips for varying levels of experience for recommended activities.
\end{itemize}
\end{tcolorbox}
\end{figure*}

\begin{figure*}[h]
\begin{tcolorbox}[title=Curated requirements for \texttt{product-gen}, colback=blue!5!white, colframe=blue!75!black, fonttitle=\bfseries]
\begin{itemize}[leftmargin=1.5em]
\item The output must highlight the product's features.
\item The output must be written in English.
\item The output must describe any features listed within the given Context in more detail.
\item The output must be at least 500 characters long.
\item The output should preferably be at least 1000 characters long.
\item The output must not use Markdown syntax.
\item The output must avoid special characters as much as possible.
\item The output must avoid excessive use of technical jargon, ensuring that the description is understandable to a general audience.
\item The output must use engaging and vivid language to capture and retain the reader's attention.
\item The output must include at least three benefits that the product provides to the user.
\item The output must follow a coherent structure, ensuring logical flow from introduction to conclusion.
\item The output must avoid any explicit comparisons with products from brands unless specified in the context.
\item The output must ensure that any numerical values or ranges are accurately represented if mentioned at all.
\item The output must include a mention of the package content.
\item The output should clearly mention any customer support or warranty information included with the product.
\item The product description should mention any personalization options available, including any important limitations or specifications.
\item The output must paint a vivid picture of the customer experience with practical use cases.
\item The output should break down complex information into clearer, more concise points.
\item The output must be free from any promotional prompts such as 'click add to cart'.
\item The output must ensure that key product information is easily skimmable.
\end{itemize}
\end{tcolorbox}
\end{figure*}

\begin{figure*}[h]
\begin{tcolorbox}[title=Curated requirements for \texttt{code-explain}, colback=blue!5!white, colframe=blue!75!black, fonttitle=\bfseries]
\begin{itemize}[leftmargin=1.5em]
\item The output should break down the code's functionality.
\item The output should explain the purpose of the code.
\item The output should use analogies and examples to clarify the explanation.
\item If technical jargon is used, the output should provide clear explanations for it.
\item The output should aim to make the explanation accessible to someone with minimal coding knowledge.
\item The output should identify and explain any variables or data structures used in the code snippet.
\item The output should detect and describe any dependencies or libraries required by the code snippet.
\item The output should check and explain any potential side effects or state changes that occur during code execution.
\item The output should include a precise, step-by-step execution order that aligns with the code.
\item If there are error handling mechanisms, the output should accurately describe them and explain how they handle potential errors.
\item The output should mention any missing components or aspects in the provided code snippet, such as lack of functionality or completeness.
\item The output should explain scenarios where certain features of the code are particularly beneficial or efficient.
\item The explanation should include potential applications and implications of the coded algorithm.
\item The output should address potential edge cases tested by the code.
\item The output should explicitly define the scope of explanation without making assumptions about specific use cases.
\item The output should not exceed 500 words to maintain conciseness and focus.
\item The output should not describe components or operations not present in the provided code.
\item The output should provide a high-level summary at the beginning to set the context.
\item The output should provide an example of how at least one function, class, or constant imported from the code can be used.
\item The output should include information about verifying the setup or configuration before execution, if applicable.
\end{itemize}
\end{tcolorbox}
\end{figure*}

\begin{figure*}[h]
\begin{tcolorbox}[title=Curated requirements for \texttt{health-consulting}, colback=blue!5!white, colframe=blue!75!black, fonttitle=\bfseries]
\begin{itemize}[leftmargin=1.5em]
\item The output should provide a concise summary of risk factors associated with health recommendations, if applicable.
\item The output should offer guidance on maintaining a healthy lifestyle, including diet and exercise recommendations, if applicable.
\item The output should prioritize listing the most critical conditions to rule out first during diagnosis, if applicable..
\item The output should provide basic first aid instructions for common injuries and illnesses, if applicable.
\item The output should encourage users to consult healthcare professionals for accurate medical assessments.
\item The output should provide action steps when experiencing severe symptoms such as difficulty breathing or swelling, if applicable..
\item The output should include important safety and handling tips for storing and managing medications, such as tracking usage duration for time-sensitive drugs, if applicable..
\item The output should recommend specific follow-up actions and time frames for reevaluation, if applicable.
\item The output should direct users to trusted medical resources and reputable websites.
\item The output should provide separate advice compartments for interventions feasible at home and those requiring professional medical environments, if applicable..
\item The output should mention simple lifestyle adjustments or practices that may help alleviate symptoms until professional care is received, if applicable.
\item The output should respond to user inquiries in clear and informative language.
\item The output should emphasize that it is not a substitute for professional medical advice, diagnosis, or treatment.
\item The output should flag symptoms that typically warrant immediate professional medical intervention, if applicable..
\item The output should provide possible explanations and suggest potential conditions based on described symptoms.
\item The output should provide specific criteria for when to transition from home care to professional medical evaluation, if applicable.
\item The output should provide information on over-the-counter and prescription medications, including side effects and dosages, if applicable..
\item The output should provide step-by-step instructions for emergency response procedures like CPR and choking, if applicable..
\item The output should include a disclaimer stating it is not a licensed medical practitioner.
\item The output should offer easy-to-understand explanations of medical terms, procedures, and concepts.
\end{itemize}
\end{tcolorbox}
\end{figure*}

\begin{figure*}[h]
\begin{tcolorbox}[title=Curated requirements for \texttt{lesson-planning}, colback=blue!5!white, colframe=blue!75!black, fonttitle=\bfseries]
\begin{itemize}[leftmargin=1.5em]
\item The lesson plan should explicitly mention extension activities for further student engagement or enrichment.
\item The lesson plan should include a detailed outline.
\item The lesson plan should be easy to follow.
\item The lesson objectives should be aligned with relevant educational standards.
\item Each section should specify the resources used.
\item The lesson plan should be well-organized.
\item The lesson plan should identify prerequisites or prior knowledge that students need before the lesson begins.
\item The lesson plan should mention strategies for engaging students in a virtual or hybrid learning environment, if applicable.
\item The outline should be broken down into an introduction, main activities, and a conclusion.
\item The lesson plan should promote critical thinking.
\item Differentiation strategies should be included and tailored to address varying student needs and abilities.
\item The lesson plan should include assessment methods to evaluate students' understanding and mastery of lesson objectives.
\item The lesson plan should cater to a specific grade level or age group.
\item The lesson plan should promote active learning.
\item The lesson objectives should be clearly stated.
\item The lesson plan should specify alternative plans if any technology used in the lesson malfunctions.
\item Each section should describe the teaching methods used.
\item Each section should describe the learning activities planned.
\item The lesson plan should be designed for a 60-minute class session.
\item The lesson objectives should be measurable.
\end{itemize}
\end{tcolorbox}
\end{figure*}

\clearpage
\subsection{Human validation of requirement validators}
\label{sec:appendix-validation}
\textbf{Validator construction.}
We employ a two-step procedure to construct reqiorement validators: 
First, we have a planner (\texttt{o3-mini}) to draft a step-by-step evaluation plan or a Python script, for each requirement.
Next, a validator either executes the Python function, or the natural language evaluation plan (with \texttt{gpt-4.1-mini}) on each example to produce the final judgment.
We iterate the validators for each requirement on the train split to achieve high human-LLM agreement.

\textbf{Human validation.}
To assess the reliability of our LLM-based requirement evaluators, we curated a set of 
1,095 evaluation results for human validation.
The evaluation results are curated with stratified sampling: 
We sample 20 evaluation results (10 positive, 10 negative) per requirement. 
If there are not enough results (e.g., when a requirement is almost always satisfied), we take the maximum number available.

We then have a human annotator manually review the evaluations. 
For each model output, the annotator compared the predicted label against their own judgment of whether the output satisfied the given requirement. 
Overall, the evaluators have 95.6\% (SD=0.08) agreement rates, indicating a reasonably high level of human–LLM agreement.

\subsection{Prompt templates and construction}
\label{sec:appendix-prompts}

We use a simple prompt template to construct our experiment prompts. The task descriptions are one-line minimal descriptions as shown in Appendix~\ref{sec:appendix-task}.

\begin{tcolorbox}[title=Prompt template for our experiments, colback=blue!5!white, colframe=blue!75!black, fonttitle=\bfseries, label={box:prompt-template}]
[Task description]\\
\\
Follow the guideline below:\\
- [requirement 1]\\
- [requirement 2]\\
...\\
- [requirement N]
\end{tcolorbox}

\begin{figure}
    \centering
    \includegraphics[width=0.4\linewidth]{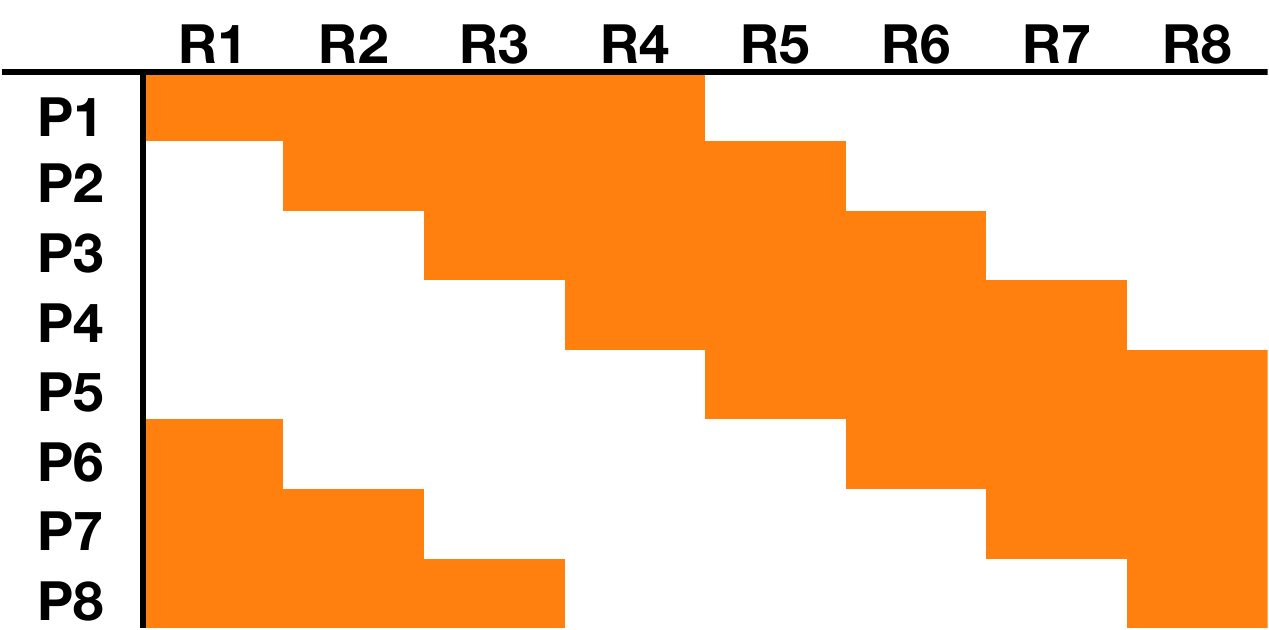}
    \caption{We use a cyclic design to generate prompts. Each prompt (row) covers the same number of $k$ consecutive requirements (column). Each requirement is specified $k$ times and unspecified $N-k$ times exactly. We randomize the order of requirements to distribute requirements from different sources.}
    \label{fig:prompt-gen}
\end{figure}

In our experiments, our goal is to systematically cover requirement subsets, to make sure (a) each prompt has roughly the same complexity in terms of the number of requirements to follow, and (b) different requirements are specified or unspecified the same number of times. 
We use a simple cyclic design to achieve this, as shown in Figure~\ref{fig:prompt-gen}.

\subsection{LLM judge setups for generic prompt optimizers}
For generic prompt optimizers, we used a requirement-agnostic LLM judge.
The judge is provided with task-level information, prompt instruction, as well as inputs and outputs from the task. 
The judge then rates the output on a 1-to-10 scale. 
We use \texttt{gpt-4.1-mini} as the judge with a temperature of 0 to ensure judge consistency, and verified that this model provides sufficiently consistent and reasonable judgments for our tasks.

\subsection{Compute resources used in the experiments}
\label{sec:appendix-compute}
In our first set of experiments (up to Section~\ref{sec:instability}), we make 6k inferences with each of the 7 models to obtain model outputs. We then make 840k inferences to evaluate the results with \texttt{gpt-4.1-mini}.

In our second experiment (Section~\ref{sec:manyreqs}), we make 6k inferences with each of the two models on 3 different prompt configurations (different numbers of requirements). We then make 720k inferences to evaluate the results with \texttt{gpt-4.1-mini}.

For prompt optimization experiments (Section~\ref{sec:fixing}), we make 16.2k inferences with each prompt optimizer to produce all optimized prompts (324k inferences for evaluation).
We then make 6k inferences with 4 different optimization results each (480k inferences for evaluation).

For LLM validators, we used a temperature of 0 to extract the most likely prediction. For other inference calls, we used the recommended temperature in best practices (1.0 for \texttt{gpt-4o} models and 0.6 for \texttt{Llama-3} models.) The maximum tokens are set to 4096, and all other inference parameters are set to default values.

\section{Detailed Analysis on Requirements Discovery}
\label{sec:appendix-discover}

\begin{figure}[t]
    \centering
    \includegraphics[width=0.8\linewidth]{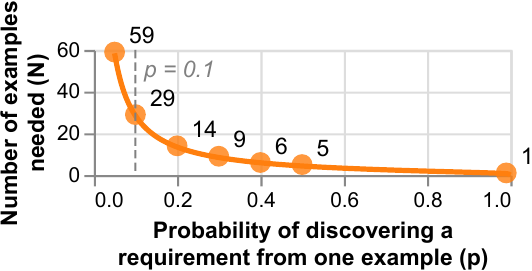}
    \caption{To discover an unspecified requirement reliably with 95\% probability, developers need to inspect a lot more examples ($N\uparrow$) when the requirement appears less frequently, gets violated less, or is harder to detect ($p\downarrow$).}
    \label{fig:prob-N}
\end{figure}

If underspecified prompts are more unstable, could developers discover \textit{relevant} task requirements in the first place? 
We argue that this is rather challenging with current manual trial-and-error prompt engineering practices,
where developers examine examples in an ad-hoc fashion and iterate their prompts only when they observe outputs that violate their expectations~\cite{liang2024prompts}.

First, many requirements are \textit{conditional} and can easily be missed when developers only look at a few representative examples -- for example, \textit{``accurate numerical values in summaries''} is only relevant to inputs containing numerical values.
When the probability of encountering such inputs ($p_\text{relevant}$) is low, the requirement is likely not to be covered within a few inspections.
However, our previous analysis demonstrates that conditional requirements are exactly where LLMs struggle more.

Second, some requirements may be violated less frequently ($p_{\text{violated}}$), and thus less likely to be discovered via observing violations. 
Yet they can still be critical, such as high-stakes safety requirements for trip advisory -- e.g., \textit{``no dangerous activities suggested.''}
In other cases, violations may be harder to recognize, with a lower probability of being noticed ($p_{\text{noticed}}$), as in \textit{``ensuring correct program execution in code explanations.''}
Moreover, when developers inspect LLM outputs, their assessments are biased by prior knowledge and subjective interpretation, which can lead them to overlook certain types of requirements~\cite{szymanski2024comparing}.

Considering these factors, an unspecified requirement may require significant efforts or luck to be discovered with the current practice. 
Quantitatively, 
a developer will need to look at $N = \log(1 - p_s) / \log(1 - p_\text{relevant} \cdot p_\text{violated} \cdot p_\text{noticed})$ examples to discover the requirement with probability $p_s$ (e.g., 0.95), assuming independent Bernoulli trials.
For example, a \textit{conditional} requirement that is relevant to 20\% of examples,  violated 50\% of the time, and perfectly noticeable will require inspecting more than 29 examples to be detected with 95\% probability (Figure~\ref{fig:prob-N}).
This will be an excessive workload for a human to complete on their own for every single model update or prompt change,
assuming they have access to a diverse dataset, if at all.

\section{Additional experiment results}

\begin{figure}[htbp]
    \centering
    \includegraphics[width=\linewidth]{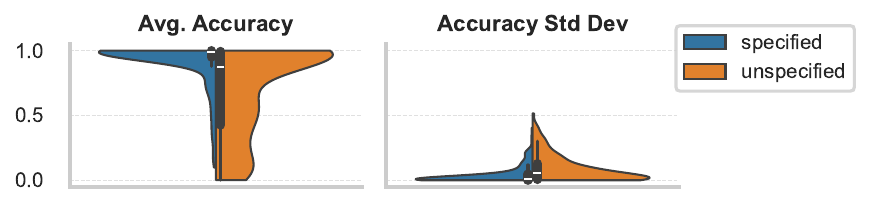}
    \caption{Comparing LLM+Prompts performances on specified requirements vs. unspecified requirements, we found that, overall, LLM+Prompts perform worse and diverge more for unspecified requirements. This is statistically significant even if we consider all other factors and explains a large portion of the variances observed (Table~\ref{tab:anova}).}
    \label{fig:overall}
\end{figure}

\begin{figure}[htbp]
    \centering
    \includegraphics[width=\linewidth]{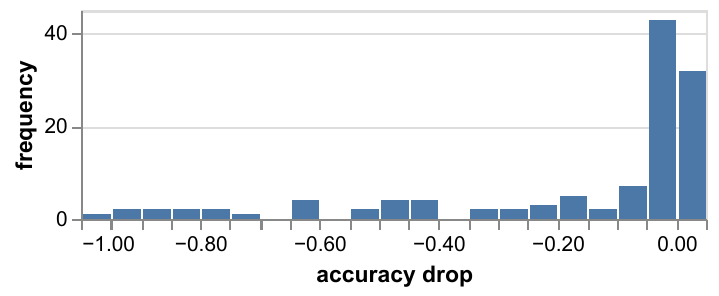}
    \caption{The histogram of average requirement accuracy drops when the prompts include more requirements (N=1$\rightarrow$19). We found 37.5\% requirements drop significantly by more than 5\%.}
    \label{fig:acc-drop-N}
\end{figure}

\subsection{Requirement guess rates breakdown}
We break down how often LLMs can ``guess'' requirements and satisfy them without specification (Table~\ref{tab:satisfaction-break-down}).
We found that LLMs excel at global, format-related, and prompt-sourced requirements, but struggle with conditional, style-related, and bottom-up requirements.

\begin{table}[h]
\centering
\begin{tabular}{lc}
\toprule
& \textbf{Fill-in Rate} \\
\midrule
\textbf{\textit{Requirement Scope}} & \\
Conditional & 22.9\% \\
Global      & 53.2\% \\
\midrule
\textbf{\textit{Requirement Source}} & \\
Bottom-up     & 21.5\% \\
Prompt-sourced     & 65.2\% \\
Top-down & 40.0\% \\
\midrule
\textbf{\textit{Requirement Category}} & \\
Content & 39.9\% \\
Format  & 70.7\% \\
Style   & 32.0\% \\
\midrule
Average   & 41.1\% \\
\bottomrule
\end{tabular}
\caption{Requirement guess rates breakdowns.}
\label{tab:satisfaction-break-down}
\end{table}

\subsection{ANOVA results}
We apply ANOVA to analyze how factors like requirement \textit{scope}, \textit{source}, \textit{category}, or \textit{model} impact LLM+Prompts performance on specified vs. unspecified requirements.
For factors that are significant, we use Tukey’s HSD test to identify which specific group means are significantly different from each other.
Results are reported in Table~\ref{tab:anova}.

\begin{table*}[h]
\centering
\begin{tabular}{r|p{2cm} p{2cm} p{1.5cm} p{1.2cm}}
\toprule
    & \textbf{Avg. accuracy} & \textbf{SD of accuracy} & \textbf{Acc. delta} & \textbf{SD delta}  \\
        \midrule
        \textbf{Specified?}   & 125.87***  & 48.47***  &  -  & -        \\ 
        \textbf{Conditional?} & 19.01*** & 16.38*** & 4.55* & 3.03     \\ 
        \textbf{Source}  & 19.99*** & 36.83*** & 9.07*** & 3.63*      \\ 
        \textbf{Category}  & 23.86*** & 55.00*** & 1.01 & 0.44      \\ 
        \textbf{Model} & 1.80 & 0.57 & 0.68 & 1.71      \\ 
        \bottomrule
      \multicolumn{2}{l}{} & \multicolumn{3}{r}{{$^{***}p<0.001,\quad ^{**}p<0.01,\quad ^{*}p<0.05$}}
\end{tabular}
\caption{ANOVA results: We report the F-value and p-value, which quantify the extent to which each variable accounts for the observed variances. 
We found that whether a requirement is specified has the largest impact on average accuracy (+0.2) and a significant impact on SD (-0.037).
Breaking down the requirements, we found models struggle with conditional requirements (-0.09 accuracy, +0.017 SD), but are better at requirements found in existing prompts and format-related requirements.
}
\label{tab:anova}
\end{table*}

\subsection{Requirements Conflict Analysis}
\label{sec:appendix-conflicts}
When we analyze the variances in \textit{specified} requirements, we observe that they are partially caused by requirement conflicts (e.g., making product descriptions more skimmable conflicts with other formatting requirements).

To better understand which requirements conflict, we conducted an analysis on requirement behaviors: 
For each requirement, we calculated its average accuracy \textit{conditioned} on whether another requirement was specified alongside it or left unspecified. 
We found that approximately 11.4\% of requirement pairs exhibited clear conflicts, defined as cases where specifying one requirement caused the accuracy of another to drop by more than 5\%. 
For example, the requirement \textit{``accessible to someone with minimal coding knowledge''} conflicted with others such as \textit{``describe error handling''} or \textit{``describe any dependencies or libraries,''} with drops in accuracy of up to 41.1\%.

Next, we assess how conflicts impact our results on requirement variances across prompts.
To assess this, we excluded all requirements showing any signs of conflict (i.e., cases where specifying one requirement reduced the accuracy of another by more than 5\%) and recomputed the variance in accuracy across prompts.

After removing these conflicting requirements, \textbf{we found the difference in prompt variance became even more pronounced}: 0.8\% for specified requirements versus 7.1\% for unspecified ones. 
This suggests that prompt instability on specified requirements often stems from conflicting requirements being included in the same prompt. 
In contrast, instability on unspecified requirements remains high even when no conflicts are present, highlighting that models are inconsistent in how they fill in missing constraints. 
For instance, the requirement \textit{``explicitly state safety guidelines specific to solo traveling''} exhibited large variance across prompts when left unspecified, even though it did not conflict with other requirements in the task. 
This indicates that the instability arises from underspecification itself, rather than requirement interaction.

\subsection{Budget-controlled comparison between CORPO and CORPO-R}

\textbf{Setups.}
We ran an experiment comparing CORPO and CORPO-R, controlling for the budget (in USD). 
We used litellm to track the usage and calculate the cost. 
We ran COPRO for 60 iterations to match the cost of COPRO-R with 9 iterations (~\$5 per run). 
We compared their results on the \texttt{code-explain} task across 20 prompts (~\$100).

\textbf{Results.}
Overall, we found that COPRO optimizers improve with more iterations (+3.6\%), only slightly falling short compared to COPRO-R (-0.2\%). 
The optimization, however, does take 2.8x longer end-to-end, as more evaluations can be easily parallelized for COPRO-R but not sequential prompt exploration in COPRO. 
We believe it is a promising future direction to explore how to best allocate resources across (more) prompt exploration and (denser) evaluation for prompt optimization.

\subsection{Optimizer robustness to training set}

\textbf{Setups.}
We ran an experiment with a different training subset to understand our Bayesian optimizer's robustness. 
For each prompt, we compute the Jaccard similarity between the requirement sets the optimizer selects under each training subset -- a score of 1 means identical selections, 0 means no overlap. 
We produced the results on the \texttt{code-explain} task across 20 prompts (cost ~\$100).

\textbf{Results.}
Overall, we found that the selected constraints are robust to training set composition, with a mean Jaccard similarity of 0.754 across all matched prompts. 
Both training sets converge on the same core constraints -- step-by-step execution order, error handling, and dependency detection consistently rank among the most-selected.

\subsection{Prompt Optimization Examples}

\begin{figure*}[ht]
\begin{tcolorbox}[title=An example of unoptimized prompts, colback=blue!5!white, colframe=blue!75!black, fonttitle=\bfseries]
Explain the code snippet.\\
\\
Follow the guideline below:\\
- The output should explain the purpose of the code.\\
- The output should explain scenarios where certain features of the code are particularly beneficial or efficient.\\
- The output should use analogies and examples to clarify the explanation.\\
- The output should not describe components or operations not present in the provided code.\\
- The output should check and explain any potential side effects or state changes that occur during code execution.\\
- The output should include a precise, step-by-step execution order that aligns with the code.\\
- The output should break down the code's functionality.\\
- If technical jargon is used, the output should provide clear explanations for it.\\
- The output should include information about verifying the setup or configuration before execution, if applicable.\\
- The output should identify and explain any variables or data structures used in the code snippet.\\
- The output should explicitly define the scope of explanation without making assumptions about specific use cases.\\
- The output should aim to make the explanation accessible to someone with minimal coding knowledge.\\
- If there are error handling mechanisms, the output should accurately describe them and explain how they handle potential errors.\\
- The output should address potential edge cases tested by the code.\\
- The output should not exceed 500 words to maintain conciseness and focus.\\
- The output should provide an example of how at least one function, class, or constant imported from the code can be used.\\
- The output should detect and describe any dependencies or libraries required by the code snippet.\\
- The explanation should include potential applications and implications of the coded algorithm.\\
- The output should provide a high-level summary at the beginning to set the context.\\
\end{tcolorbox}
    \caption{Unoptimized prompts generated from the prompt template (acc=75.4\%).}
    \label{fig:unoptimized-prompt}
\end{figure*}

\begin{figure*}
\begin{tcolorbox}[title=An example of COPRO-optimized prompts, colback=blue!5!white, colframe=blue!75!black, fonttitle=\bfseries]
Provide a detailed explanation of the given code snippet, following these guidelines:\\
\\
1. **Overview**: Begin with a high-level summary that sets the context for the code's purpose and functionality.\\
2. **Purpose and Benefits**: Clearly explain the code's main objective and highlight scenarios where its features are particularly beneficial or efficient.\\
3. **Analogies and Examples**: Use relatable analogies and examples to enhance understanding, especially for complex concepts.\\
4. **Code Breakdown**: Decompose the code into its fundamental components, explaining the role and function of each part.\\
5. **Step-by-Step Execution**: Offer a precise, sequential walkthrough of how the code executes, ensuring clarity on the process.\\
6. **State Changes and Side Effects**: Identify any potential side effects or state changes that occur during execution.\\
7. **Variable and Data Structure Explanation**: Define and explain any variables or data structures used, ensuring comprehension for those with minimal coding knowledge.\\
8. **Technical Jargon**: Simplify or clarify any technical terms to make the explanation accessible.\\
9. **Setup Verification**: Include information on verifying the setup or configuration before running the code, if relevant.\\
10. **Error Handling**: Describe any error handling mechanisms, detailing how they manage potential errors and edge cases.\\
11. **Dependencies**: Identify any libraries or dependencies required by the code.\\
12. **Applications and Implications**: Discuss potential applications and implications of the algorithm or functionality provided by the code.\\
13. **Function Usage Example**: Provide an example of how at least one function, class, or constant from the code can be utilized.\\
14. **Conciseness**: Ensure the explanation does not exceed 500 words, maintaining focus and clarity. 
\end{tcolorbox}
    \caption{COPRO-optimized prompts (acc=86.7\%). We found COPRO-optimized prompts tend to reorder requirements in a more logical structure, merge related requirements together, and sometimes drop requirements.}
    \label{fig:optimized-prompt}
\end{figure*}

\clearpage

\section{Additional tasks setups and results}
\label{sec:appendix-additional-tasks}
\paragraph{Tasks and data.}
We analyzed two more tasks, \texttt{health-consulting} and \texttt{lesson-planning}, from different categories (Physical Sciences, Education) in the Anthropic report.

\begin{itemize}[noitemsep, topsep=0pt]
    \item \texttt{health-consulting}: Offer symptom guidance, medical info, first-aid tips, and health advice.
    \item \texttt{lesson-planning}: Generate a lesson plan for the given user query.
\end{itemize}

We used two existing datasets to run the LLM+Prompts on: 
HealthBench~\cite{arora2025healthbench} for health consulting,
and lesson\_plan~\cite{samadeniyi2026lessonplan} for lesson planning.

\begin{itemize}[noitemsep, topsep=0pt]
 \item For HealthBench, we use \texttt{gpt-4.1-mini} to label if each query is a medical consulting request from a user. We filter out the ones that are not medical consulting and the ones that are not in English. We then sample 200 examples from the dataset.

 \item For lesson\_plan, we drop duplicated queries and then sample 200 examples from the dataset.
\end{itemize}

\paragraph{Requirements.}
We followed the same procedure to curate requirements as in Appendix~\ref{sec:appendix-elicitation} and shared them in Appendix~\ref{sec:appendix-requirements}.
We found existing prompts provided by Anthropic and GPTs\footnote{\url{https://platform.claude.com/docs/en/resources/prompt-library/lesson-planner},\\\url{https://github.com/linexjlin/GPTs/blob/3adfb7b38423b64a995057483c1f9007ed5f4da5/prompts/Medical Diagnosis Assistant.md}}.

\paragraph{Results.}
We observed very similar empirical results in the additional tasks we studied, showing our main findings can be readily generalized to different tasks and models: 

\begin{enumerate}[label=(\arabic*), noitemsep, topsep=0pt]
    \item There is a performance gap between specified and unspecified requirements (83.6\% vs. 59.5\%); models are able to guess unspecified requirements frequently (25.1\%); format-related / global / developer-written requirements are easier to guess (cf. Section~\ref{sec:default-behaviors}).
    \item Unspecified requirements are more likely to regress over model updates (4.3\% vs. 2.3\%) (cf. Section~\ref{sec:instability}).
    \item LLMs are struggling with more requirements (dropping from 95.0\% to 78.5\% when we increase the number from 1 to 19) (cf. Section~\ref{sec:manyreqs}).
    \item We also found the results hold for newer reasoning models we tested (\texttt{gemini-2.5-flash}).
\end{enumerate}

\section{Details on Agentic Coding Case Study}
\label{sec:appendix-case-study}

\paragraph{Prompt and requirements.}
We identified a system prompt for front-end development from Cursor Community.\footnote{\url{https://cursor.directory/gatsby-development-best-practices}}
The system prompt consists of a list of 25 requirements grouped by categories (e.g., TypeScript Usage, UI and Styling).

\paragraph{Optimizers.}
We use the same optimizers with the same setup as in Section~\ref{sec:optimizer}.
For the Bayesian optimizer, we split the prompt into different segments based on the requirements, and the optimizer will choose the segments to include in the final prompt.

\paragraph{Evaluation.}
For a proposed prompt, we run it on each user query three times to reduce the variance of the generated code repository. 
Each run is executed with OpenHands scaffolds using \texttt{Qwen3-30B-A3B-Instruct-2507}, provided with TerminalTool, FileEditorTool, and TaskTrackerTool.
The generated code repository will then be packed into one single file with Repomix and provided to \texttt{gemini-2.5-flash} for evaluation.
The evaluator model has access to all 25 requirements and will evaluate the code repository based on how well it adheres to the requirements, producing an average accuracy across requirements.

\begin{figure*}[ht]
\begin{tcolorbox}[title=An example of unoptimized prompts, colback=blue!5!white, colframe=blue!75!black, fonttitle=\bfseries]
You are an expert in TypeScript, Gatsby, React and Tailwind.\\

Code Style and Structure

- Write concise, technical TypeScript code.\\
- Use functional and declarative programming patterns; avoid classes.\\
- Prefer iteration and modularization over code duplication.\\
- Use descriptive variable names with auxiliary verbs (e.g., isLoaded, hasError).\\
- Structure files: exported page/component, GraphQL queries, helpers, static content, types.\\

Naming Conventions

- Favor named exports for components and utilities.\\
- Prefix GraphQL query files with use (e.g., useSiteMetadata.ts).\\

TypeScript Usage

- Use TypeScript for all code; prefer interfaces over types.\\
- Avoid enums; use objects or maps instead.\\
- Avoid using `any` or `unknown` unless absolutely necessary. Look for type definitions in the codebase instead.\\
- Avoid type assertions with `as` or `!`.\\

Syntax and Formatting

- Use the "function" keyword for pure functions.\\
- Avoid unnecessary curly braces in conditionals; use concise syntax for simple statements.\\
- Use declarative JSX, keeping JSX minimal and readable.\\

UI and Styling

- Use Tailwind for utility-based styling\\
- Use a mobile-first approach\\

Gatsby Best Practices

- Use Gatsby's useStaticQuery for querying GraphQL data at build time.\\
- Use gatsby-node.js for programmatically creating pages based on static data.\\
- Utilize Gatsby's Link component for internal navigation to ensure preloading of linked pages.\\
- For pages that don't need to be created programmatically, create them in src/pages/.\\
- Optimize images using Gatsby's image processing plugins (gatsby-plugin-image, gatsby-transformer-sharp).\\
- Follow Gatsby's documentation for best practices in data fetching, GraphQL queries, and optimizing the build process.\\
- Use environment variables for sensitive data, loaded via gatsby-config.js.\\
- Utilize gatsby-browser.js and gatsby-ssr.js for handling browser and SSR-specific APIs.\\
- Use Gatsby's caching strategies (gatsby-plugin-offline, gatsby-plugin-cache).

Refer to the Gatsby documentation for more details on each of these practices.
\end{tcolorbox}
    \caption{Unoptimized prompt from Cursor Community (acc=44.1\%).}
    \label{fig:unoptimized-prompt-codegen}
\end{figure*}

\begin{figure*}
\begin{tcolorbox}[title=An example of Bayesian-optimized prompts, colback=blue!5!white, colframe=blue!75!black, fonttitle=\bfseries]
You are an expert in TypeScript, Gatsby, React and Tailwind.

- Prefer iteration and modularization over code duplication.\\
- Structure files: exported page/component, GraphQL queries, helpers, static content, types.\\

- Prefix GraphQL query files with use (e.g., useSiteMetadata.ts).\\

TypeScript Usage

- Use TypeScript for all code; prefer interfaces over types.\\
- Avoid enums; use objects or maps instead.\\

UI and Styling

- Use Tailwind for utility-based styling\\
- Use a mobile-first approach\\

- Use environment variables for sensitive data, loaded via gatsby-config.js.
\end{tcolorbox}
    \caption{Bayesian-optimized prompts (acc=47.4\%). We found Bayesian-optimized prompts use much fewer tokens while achieving similar accuracies.}
    \label{fig:optimized-prompt-codegen}
\end{figure*}

\end{document}